\documentclass[letterpaper, 10 pt, conference]{ieeeconf}  

\IEEEoverridecommandlockouts                              

\usepackage{graphics} 
\usepackage{epsfig} 
\usepackage{mathptmx} 
\usepackage{mathrsfs}
\usepackage{times} 
\usepackage{amsmath} 

\usepackage{amsthm}
\usepackage{amssymb}  
\usepackage{wasysym}
\usepackage{txfonts}
\usepackage{bbm}
\usepackage{color}  
\usepackage[dvipsnames]{xcolor}
\usepackage{indentfirst}
\usepackage{multirow}

\usepackage{color, colortbl}
\usepackage{algorithm}
\usepackage{algpseudocode}

\usepackage{tikz}
\usetikzlibrary{arrows,shapes,trees,calc}
\usetikzlibrary{backgrounds}
\usepackage{pgfplots}
\pgfplotsset{compat=1.14}
\usepgfplotslibrary{polar}
\usepgfplotslibrary{patchplots}

\newcolumntype{P}[1]{>{\centering\arraybackslash}p{#1}}
\usepackage{boldline}
 \usepackage{textcomp}
\newcounter{RamCount}
\newcounter{ShannonCount}
\newcounter{PatrickCount}

\addtocounter{ShannonCount}{1}
\addtocounter{PatrickCount}{1}
\addtocounter{RamCount}{1}

\newcommand{\shan}[1]{\textcolor{WildStrawberry}{\textbf{(\theShannonCount) Shannon}: (#1)}\addtocounter{ShannonCount}{1}}
\newcommand{\pat}[1]{\textcolor{OliveGreen}{\textbf{(\thePatrickCount) Patrick}: (#1)}\addtocounter{PatrickCount}{1}}
\newcommand{\ram}[1]{\textcolor{CornflowerBlue}{\textbf{(\theRamCount) Ram}: (#1)}\addtocounter{RamCount}{1}}

\usepackage[
    style=ieee,
    doi=false,
    isbn=false,
    url=false,
    eprint=false,
    backend=bibtex,
    natbib=true
    ]{biblatex}

\bibliography{references}

\include{commands}
\usepackage{safe_prosthesis_notation}

\usepackage{hyperref}

\title{\LARGE \bf Trip Recovery in Lower-Limb Prostheses using Reachable Sets of Predicted Human Motion}
\author{Shannon M. Danforth, Patrick D. Holmes, and Ram Vasudevan 
\thanks{This material is based on work supported by the National Science Foundation Career Award \#1751093 and and by the Office of Naval Research under Award Number N00014-18-1-2575.}%
\thanks{S.M. Danforth, P.D. Holmes, and R. Vasudevan are with the Department of Mechanical Engineering at the University of Michigan,
        Ann Arbor, MI, USA.
        {\tt\small \{sdanfort, pdholmes, ramv\} @umich.edu}}%
}

\begin{document}
\setlength{\textfloatsep}{8pt}

\maketitle
\thispagestyle{empty}
\pagestyle{empty}

\begin{abstract}
People with lower-limb loss, the majority of which use passive prostheses, exhibit a high incidence of falls each year. 
Powered lower-limb prostheses have the potential to reduce fall rates by actively helping the user recover from a stumble, but the unpredictability of the human response makes it difficult to design controllers that ensure a successful recovery. 
This paper presents a method called TRIP-RTD (Trip Recovery in Prostheses via Reachability-based Trajectory Design) for online trajectory planning in a knee prosthesis during and after a stumble that can accommodate a set of possible predictions of human behavior.
Using this predicted set of human behavior, the proposed method computes a parameterized reachable set of trajectories for the human-prosthesis system. 
To ensure safety at run-time, TRIP-RTD selects a trajectory for the prosthesis that guarantees that all possible states of the human-prosthesis system at touchdown arrive in the basin of attraction of the nominal behavior of the system.
In simulated stumble experiments where a nominal phase-based controller was unable to help the system recover, TRIP-RTD produced trajectories in under 101 ms that led to successful recoveries for all feasible solutions found.
\end{abstract}

\section{Introduction}
\label{sec:intro}

In $2005$, an estimated $623,000$ people were living with major lower-limb loss in the United States, with the number predicted to double by $2050$ \cite{ziegler-graham2008}.
The majority of these people use passive lower-limb prostheses, with instances of falling comparable to older adults over $85$ \cite{miller2001, blake1988}.
Powered lower-limb prostheses have the potential to lower metabolic cost and improve gait symmetry \cite{wong2015}, and recent developments in their control allow them to assist users while walking up slopes, ascending/descending stairs, and walking backwards \cite{tucker2015}.
However, despite having the potential to assist users during slips or trips, most powered prostheses do not include stumble-recovery features.
Equipping powered prostheses with stumble-recovery features could lead to fewer falls and increased balance confidence in lower-limb prosthesis users. 

Before discussing why stumble-recovery control is difficult in prostheses, it is useful to understand their general control framework.
During locomotion, robotic lower-limb prostheses use 
a high-level controller to estimate the human's intent.
Common high-level controllers include activity mode detection \cite{sup2007, huang2009},
direct volitional control \cite{FERRIS2006},
or a combination of both \cite{rezazadeh2019}.
A mid-level controller then converts this estimated intent into a desired trajectory for a low-level controller to track \cite{tucker2015}.

Two challenges contribute to the difficulty of designing effective stumble-recovery controllers within this hierarchical control framework.
The first is reliably detecting the stumble itself.
In stumble experiments, humans have exhibited responses with latencies under $100$ ms \cite{schillings1999}. 
Researchers have shown that accelerometers mounted on the prosthesis can detect stumbles within this short time span, although false positives are common \cite{lawson2010, zhang2011}.
The second challenge lies in choosing a recovery behavior for the prosthesis once the stumble has been detected. 
Difficulties in this area arise from 
the presence of uncertainty in estimates of human intent and the inability of existing mid-level controllers 
to generate trajectories for stumble recovery in real-time.

\begin{figure}
    \centering
    \includegraphics[trim={0 0 0 0},clip, width=\columnwidth]{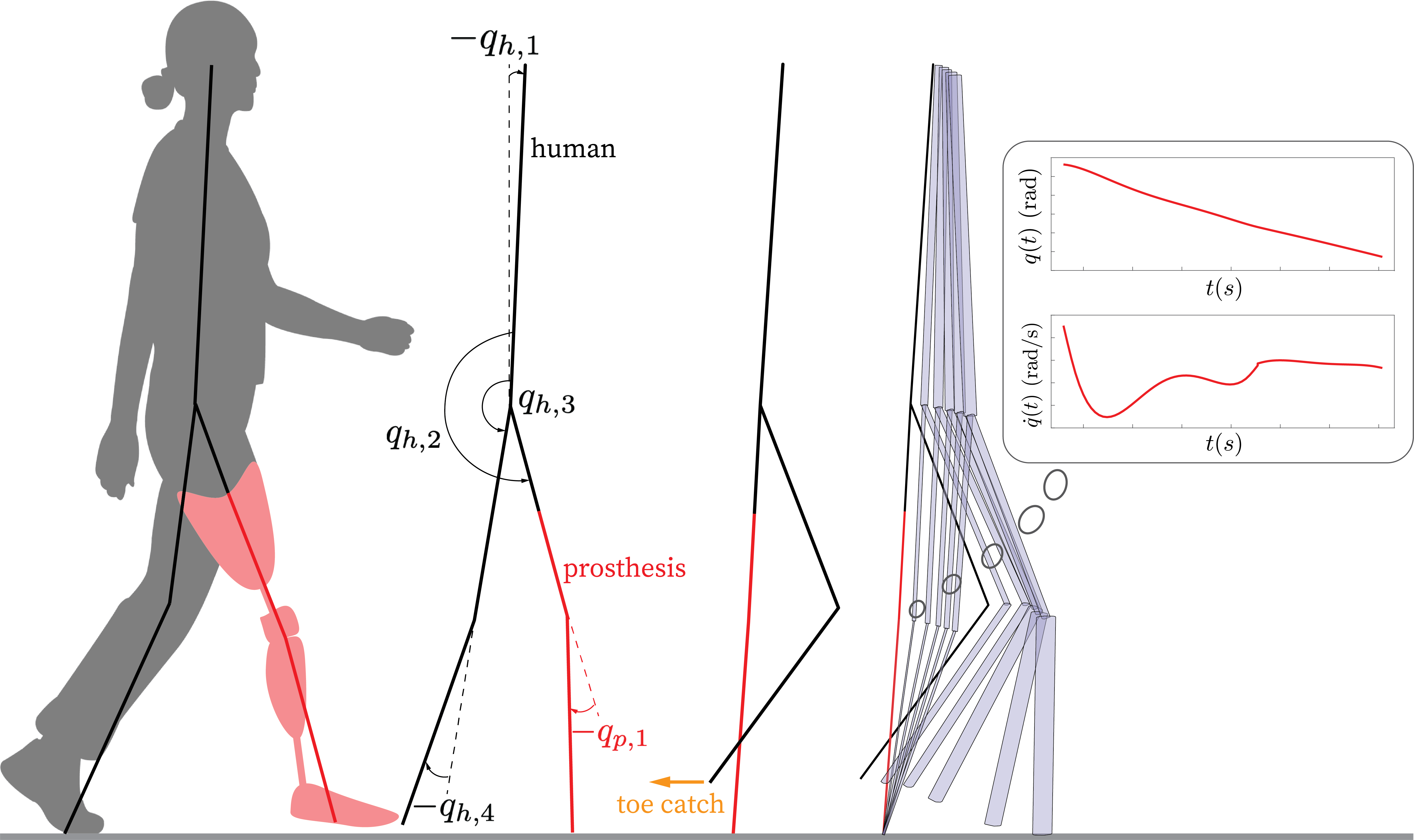}
    \caption{The TRIP-RTD framework plans trajectories for a knee prosthesis during and after a stumble that ensure successful recoveries for a set of human behavior.
    The bipedal model used in this work contains human and prosthesis subsystems, and undergoes a simulated toe catch.
    This figure shows the system's reachable sets, which depend on both human and prosthetic behavior, after a toe catch perturbation.
    TRIP-RTD generates a reference trajectory for the prosthesis that guarantees the model's successful recovery for any human behavior in a set of predictions.}
    \label{fig:overviewfig}
\end{figure}

As a human undergoes a stumble and tries to recover, 
any uncertainty in the controller's estimate of the human intent or current state could increase fall risk by producing unexpected torques \cite{tucker2015}.
This uncertainty is present in controllers that rely on training data and are at risk of misclassifying intent \cite{zhong2020} 
as well as controllers that use an estimated human state to prescribe joint torques.
Two widely-used prosthetic control methods that fall into the latter category are virtual constraint control \cite{gregg2013}
and finite-state control \cite{herr2009}.
Both methods express prosthetic target trajectories and torques as functions of gait phase instead of time, allowing for generalization across tasks, speeds, and users.
However, 
all of these control schemes are based on a single (most likely) estimate of human behavior, which could prove disastrous if this estimate is wrong. 
Instead, a more robust approach would account for uncertainty by designing controllers that are suitable across a range of possible human behaviors.
In addition to estimation uncertainty, the variation in human stumble recoveries \cite{shirota2014} makes it impossible to predict exactly how the human will respond.
Recent work has proposed a quadratic program for planning post-stumble recoveries \cite{thatte19_rss}, but in general, the high-dimensional, nonlinear, hybrid nature of the human-prosthesis system makes online optimization methods difficult.
Generating trajectories online while accounting for a range of possible human stumble-recovery behavior adds an additional challenge.
With these challenges in mind, we present a method for planning a prosthetic recovery strategy online that accounts for a range of human behavior.
We draw this idea from Reachability-based Trajectory Design (RTD), which uses reachable sets within a trajectory-planning framework to guarantee collision-free operation of autonomous robots.
\cite{kousik2018}. 
Recently, RTD was applied to a 7-DOF robotic manipulator (ARMTD), composing reachable sets of each joint in a local coordinate frame into a reachable set of the robot in the global workspace \cite{holmes20}.

In this work, we apply a similar reachability-based approach to a model of a human with a transfemoral, unilateral amputation and a robotic knee prosthesis.
We call our method TRIP-RTD (Trip Recovery in Prostheses via Reachability-based Trajectory Design).
Because studies have highlighted the support leg's critical role in trip recovery \cite{shirota2015, pijnappels2004}, this work focuses on planning trajectories for a prosthesis in stance as the human's swing leg undergoes a ``toe catch'' stumble.
We simulate a range of stumble-recovery behaviors for this model, parameterize the joint trajectories, and use them to compute forward reachable sets of the human-prosthesis system in the global coordinate frame.
Online, we use these reachable sets to select prosthetic knee behavior that ensures the system can successfully recover from the toe catch.
An overview of TRIP-RTD can be found in Fig. \ref{fig:overviewfig}.
This work does not include stumble detection or real human stumble-recovery data, and instead relies on simulated data of a bipedal robot model. 
However, the framework is generalizable to different models, perturbation types, and stumble-recovery datasets.

Our contributions are as follows. 
First, a new generalizable framework, described in Section \ref{sec:theory}, for planning prosthetic stumble-recovery trajectories online that account for a range of predicted human behavior.
Second, an implementation of the method using simulated stumble-recovery data, outlined in Section \ref{sec:implementation}.
Third, an experiment comparing the proposed method to nominal phase-based control, described in Section \ref{sec:experiment}, with results presented in Section \ref{sec:results}.
Reference trajectories generated by feasible solutions in the TRIP-RTD framework led to a successful recovery $100\%$ of the time, whereas nominal phase-based control never resulted in success.
A discussion on the method's effectiveness and incorporating human data in future work takes place in Section \ref{sec:discussion}.
\section{Theory}
\label{sec:theory}
In this section, we describe the framework TRIP-RTD uses to plan and execute safe stumble-recovery trajectories.
This section begins by introducing the model and dynamics of the human-prosthesis system.
Next, it defines a set of states at heel strike, called the target set $\XT$, from which a nominal controller can return the system to its nominal limit cycle walking behavior.
Subsequently, a model of human stumble-recovery behavior that incorporates uncertainty is constructed.
Utilizing the target set and parameterized stumble recoveries, we formalize our approach to designing safe prosthetic joint trajectories within an optimization problem.
These trajectories satisfy constraints that keep the system from falling, while minimizing a user-specified cost function.

\subsection{Human-Prosthesis Model}
\label{subsec:theory_model}
This work considers a hybrid model of the human-prosthesis system in the sagittal plane.
Let the configuration $q(t) \in Q \subset \R^{n_q}$ be a vector of $n_q$ generalized coordinates of the model at time $t$.
Let $x(t) = \begin{bmatrix} q^T(t) & \dot{q}^T(t)
    \end{bmatrix}^T  \in X \subset \R^{2n_q}$ be the state of the system at time $t$.
We consider two subsystems of this model, broken into $\qh(t)$ and $\qp(t)$ (and $\xh(t), \xp(t)$) representing the degrees of freedom at time $t$ controlled by the human and prosthesis, respectively, where $q(t) = \begin{bmatrix}
    \qh^T(t) &
    \qp^T(t)
    \end{bmatrix}^T$ and $x(t) = \begin{bmatrix}
    \xh^T(t) &
    \xp^T(t)
    \end{bmatrix}^T$.

We assume that the hybrid model has only one discrete mode representing single support, so that only one foot is on the ground at any time.
Within this mode, the dynamics can be written as \cite[Sec. 1.2]{westervelt2007}: 
\begin{equation}
    M(q(t))\ddot{q}(t) + C(q(t), \dot{q}(t))\dot{q}(t) + G(q(t)) = B(q(t))u(t)
\end{equation}
where $M$ represents the mass matrix, $C$ represents Coriolis forces, $G$ represents forces due to gravity, and $B$ maps inputs to generalized accelerations.
These dynamics may be derived via the method of Lagrange.
We may rewrite these dynamics as a first order ordinary differential equation:
\begin{equation}
    \dot{x}(t) = f(x(t)) + g(x(t))u(t).
\end{equation}
We make the following assumption about the control inputs which we are allowed to control in this model:
\begin{assum}
We assume that we may only choose the inputs to the prosthetic subsystem $\up$, and that the remaining inputs $\uh$ are determined by the human user.
\end{assum}

A reset map is used to model the change in velocity of the model due to impacts with the ground.
Let $\dot{x}^-(t)$ and $\dot{x}^+(t)$ be the pre- and post-impact states of the model, which are related by:
\begin{equation}
    x^+(t) = \Delta(x^-(t)),
\end{equation}
where the reset map $\Delta$ may be derived via conservation of angular momentum arguments \cite{uchida2015}.
Finally, let $\switching$ represent the set of states that activate a switching condition, such as impact of the swing foot with the ground.
The hybrid model may then be written as \cite{westervelt2007}:
\begin{align}
    \Sigma : \begin{cases}
         \dot{x}(t) = f(x(t)) + g(x(t))u(t) \quad &x  \not\in \switching \\
         x^+(t) = \Delta(x^-(t)) \quad &x \in \switching.
         \end{cases}
\end{align}
Throughout this paper, we assume the switching condition is activated upon the swing foot's initial contact with the ground, and refer to this event as \defemph{heel strike}. 
In human walking, heel strike initiates a double-support phase that lasts about $1/6$ of stance \cite{winter1991}; however, it is commonly modeled as an instantaneous impact in bipedal walking systems \cite{westervelt2007}. 

This work assumes that during unperturbed walking, a phase-based control scheme is used to control the prosthesis.
 As described in the introduction, phase-based techniques have been successfully applied to control prostheses in the real world, and rely on the existence of a phase variable to convert human states into target joint trajectories.
The phase variable is defined as follows: 
\begin{defn}(\cite{rezazadeh2019})
The phase variable $\theta: \R^{n_q} \to \R$ is a function only of the generalized coordinates, and is monotonically non-decreasing throughout the stance phase.
\end{defn}
\noindent An example phase variable is the stance leg angle, depicted for a specific model in Fig. \ref{fig:experiment_results}(b), which progresses monotonically throughout stance.
Usually the phase variable is evaluated as a function of a desired trajectory of the generalized coordinates that is parameterized by time, i.e. $\theta(q(t))$. 
Finally, we assume that there exists a nominal prosthesis controller:
\begin{assum}
There exists a nominal prosthesis controller $\unom(\theta)$ that generates a locally-stable nominal walking limit cycle $\xnom$.
\end{assum}
\noindent Note that such a nominal prosthesis controller can be constructed by using phase-based control \cite{gregg2013}.

\subsection{Parameterized Stumble-Recovery Behaviors}
\label{subsec:theory_parameters}

Next, we formally define stumbles.
Without loss of generality, let the current stance phase start at time $t = 0$ and end when the switching condition is satisfied at time $t = \tf$.
\begin{assum}
The stumble occurs between times $[\tsm, \tsp] \subset [0, \tf]$, is immediately detected, and the configuration $q_0 = q(\tsm)$ is known.
\end{assum}
Next, to describe a range of possible human responses to perturbations, and to design trajectories of the robotic prosthesis for control, we introduce the notion of \defemph{trajectory parameters} that describe the evolution of generalized coordinates over time.


\begin{defn}
Let $K \subset \R^{n_k}$ be a compact set of $n_k$ \defemph{trajectory parameters}, where each $k \in K$ maps to a trajectory of the generalized coordinates $\tilde{q} : [\tsm, \tf] \to Q$. 
Let $\tilde{q}(t; k)$ be the configuration at time $t \in [\tsm, \tf]$ parameterized by $k \in K$.
We assume $\dot{\tilde{q}}(\cdot; k)$ is continuous over times $t \in [\tsm, \tsp)$ and $t \in (\tsp, \tf]$, but we allow for a jump in velocity at time $t = \tsp$. WLOG, we require $\tilde{q}(\tsm; k) = 0$.
\end{defn}
\noindent We choose $\tilde{q}(\tsm; k) = 0$ so that online we can construct reachable sets that use the initial configuration $q_0$ at the start of the stumble, described in Sec. \ref{subsec:theory_reachability}.
We differentiate between trajectory parameters of the prosthesis and the human subsystems.
Let $\kp \in \Kp$ denote a trajectory parameter for the prosthesis subsystem, and $\kh \in \Kh$ denote a trajectory parameter of the human subsystem.
We let $\tqp(t; \kp)$ and $\tqh(t; \kh)$ be the configurations of each subsystem at time $t \in [\tsm, \tf]$ parameterized by $\kp \in \Kp$ and $\kh \in \Kh$, respectively, and let $k = \{\kp\} \times \{\kh\}$, $K = \Kh \times \Kp$, and $n_k = \nkh + \nkp$.
Notice that this formulation assumes that the parameterized trajectories of the prosthetic and human subsystems are \emph{independent} of one another.

We make the following assumption about estimating human trajectory parameters online:
\begin{assum}
At the onset of the stumble ($ t = \tsm$), a subset of predicted human trajectory parameters $\Khbar \subseteq \Kh$ is generated which contains the true trajectory parameter chosen by the human.
\end{assum}
The goal of TRIP-RTD is to select a trajectory parameter $\kp \in \Kp$ that yields a safe trajectory of the whole system across all $\kh \in \Khbar$.
In the next section, we introduce \defemph{target sets} that decide whether or not a trajectory is safe.

\subsection{Target Set}
\label{subsec:theory_target}

As we describe in Sec. \ref{subsec:theory_optimization}, our goal is to perform real-time optimization to generate a control input for the prosthesis that can ensure that after a stumble, an individual is able to return to nominal walking without falling. 
We formalize this as an optimization problem which tries to return an individual to the basin of attraction of the nominal prosthesis controller. 
Rather than return the individual's state to the basin of attraction at an arbitrary time, we instead define a target set $\XT$ that is a subset of the basin of attraction of the nominal prosthesis controller at heel strike:
\begin{defn}
\label{def:target_set}
The target set $\XT \subset \R^{2n_q}$ is a set of states such that if the state of the system at heel strike $x(\tf)$ lies within $\XT$, the system will eventually converge to the nominal limit cycle $\xnom$ when applying the nominal controller $\unom$ for all subsequent steps.
\end{defn}
The target set $\XT$ may be a polytope encompassing all states $x(\tf)$ that converge to the limit cycle, similar to \cite{holmes2020characterizing}.

\subsection{Reachable Sets of the Human-Prosthesis System}
\label{subsec:theory_reachability}

Because it is impossible to perfectly predict human behavior, we want to choose a desired trajectory of the prosthesis (represented by $\kp \in \Kp$) that is robust to a range of plausible human stumble-recovery responses (represented by $\Khbar \subseteq \Kh$).
In particular, Sec. \ref{subsec:theory_optimization} describes our method for choosing $\kp \in \Kp$ that brings the system to the target set $\XT$ for all $\kh \in \Khbar$.
To do so, we characterize the forward reachable set $\FRS(\cdot \, ; \, \Khbar): \Kp \to \P(\R^{2n_q})$ that maps $\kp \in \Kp$ to subsets of the state space at heel strike, given $\Khbar$:
\begin{align}
    \label{eqn:FRS_theory}
    \begin{split}
    \FRS(\kp ; \Khbar) = \bigg\{\tilde{x} &\in \P(\R^{2n_q}) \ \Big| \ \exists k \in \{k_p\} \times \Khbar \text{ s.t. } \\ & \tilde{x} = \begin{bmatrix}q_0 + \int_{\tsm}^{\tf} \dot{\tilde{q}}(\tau; k) d\tau\\ \dot{\tilde{q}}(\tf; k))\end{bmatrix} \bigg \}.
    \end{split}
\end{align}
where $\P(\R^{2n_q})$ is the power set of the state space.
Notice that the configuration at the start of the stumble $q_0$ is incorporated into the reachable set.
We show how to compute an overapproximation of this object in Sec. \ref{sec:implementation:rotatotopes}.
To be clear, in this work we use the term ``reachable sets'' to refer to sets of parameterized \emph{kinematic} trajectories.
We use these reachable sets to produce desired trajectories that the \emph{dynamic} model of the prosthesis attempts to track via a tracking controller.

\subsection{Online Trajectory Optimization}
\label{subsec:theory_optimization}
The goal of this work is to plan and execute trajectories of a prosthesis that robustly respond to stumbles that are sensed online.
Importantly, we want to choose trajectory parameters $\kp$ which lead to successful recovery across a range of plausible human trajectories, represented by the subset $\Khbar \subseteq \Kh$.
We formulate this as an optimization problem, with constraints ensuring that the reachable set $\FRS( \kp  ; \Khbar)$ corresponding to $\kp$ is contained within the target set $\XT$.
Given a $\Khbar$, let $\Kunsafe(\Khbar)$ be the set of \emph{unsafe} trajectory parameters for the prosthesis which is formally defined as follows
\begin{equation}
    \Kunsafe(\Khbar) = \{ \kp \ | \ \FRS(\kp ; \Khbar) \not\subseteq \XT \} ,
\end{equation}
so a trajectory parameter $\kp$ is unsafe if the reachable set corresponding to that parameter is not fully contained within $\XT$.
Letting $\costfunc(\kp)$ be a user-specified cost function, we solve for the optimal trajectory parameter $\kp$ as
\begin{equation}
    \kpopt = \regtext{argmin}_{\kp \in \Kp}\big\{\costfunc(\kp)\ |\ \kp \not\in \Kunsafe(\Khbar)\big\}.
\end{equation}
For example, $\costfunc(\kp)$ may penalize the distance from a desired trajectory, or the work done by the prosthesis.

Once an optimal trajectory parameter has been identified, the control input for the prosthetic knee at time $t$ is given by a tracking controller
\begin{equation} \label{eq:feedback}
    u_{p}(t) =  G_1\big( q_{p}(t) - \tilde{q}_{p}(t;k)\big) + G_2\big( \dot{q}_{p}(t) - \dot{\tilde{q}}_{p}(t;k) \big),
\end{equation}
where $G_1$ and $G_2$ are user specified feedback gains.
Next, we describe how to implement TRIP-RTD in practice.




\section{Implementation}
\label{sec:implementation}

This section describes our implementation of the approach presented in Section \ref{sec:theory}.
We develop a dynamic model of the human/prosthesis system, and simulate system responses to a toe-catch perturbation in the contralateral swing leg.
Using parameterized trajectories of these simulated responses, we create target sets $\XT$ and reachable sets for the human-prosthesis system.
We use these sets to form constraints in an online optimization problem to ensure a successful recovery across a range of possible human stumble responses.

\subsection{Human-Prosthesis Model} \label{subsec:implementation_model}
This work uses a $10$-dimensional model inspired by the robot Rabbit \cite{chevallereau2003}, altered to include both a human and prosthesis subsystem, shown in Fig. \ref{fig:overviewfig}.
The model is underactuated in single support, due to a lack of feet. 
We assume the human has a unilateral amputation and wears a transfemoral prosthesis, where the prosthetic knee joint torque is the only input to the model we control.
We assume the model utilizes phase-based control during nominal walking, where the stance leg angle $\theta$, shown in Fig. \ref{fig:experiment_results}(b), is used as the phase variable. 
The stance leg angle is given by $\theta(q(t)) = -q_{h,1}(t) - q_{h,2}(t) - \frac{q_{p,1}(t)}{2}$ and is monotonic in time.
For simplicity, we write $\theta(q(t))$ as $\theta$ throughout this paper.
During one stance phase, $\theta$ ranges from $\theta_0$ to $\theta_f$.
The system walks continuously in alternating phases of single and double support, with the double support phase being instantaneous.
Left-right symmetry in the gait is not required.

\subsection{Parameterized Stumble-Recovery Behaviors} \label{subsec:implementation_params}
We created nominal, time-varying, periodic walking trajectories for the entire system using using the trajectory optimization toolbox FROST \cite{Hereid2017}, parameterized by $\theta$.
The nominal FROST trajectories spanned $10$ different step lengths.
Before and after the perturbation/recovery, the prosthetic knee uses phase-based control, with $\theta$ mapping to nominal knee position and velocity values.
In the absence of experimental human stumble data, we simulated stumble-recovery responses.
The simulated trials described in this subsection are used to obtain the set of human stumble-recovery responses parameterized by $\Kh$ (Sec. \ref{subsec:theory_parameters}), the prosthetic knee trajectory library parameterized by $\Kp$ (Sec. \ref{subsec:theory_parameters}), and the target set of states $\XT$ (Sec. \ref{subsec:theory_target}).
The toe catch was simulated for all $10$ FROST nominal gaits.
All perturbations occurred when the prosthetic knee was in stance, and lasted from $60-80\%$ of the nominal $\theta$-range for a given step length.

During the perturbation, the swing foot (via torques applied to the hip and swing knee) was driven to remain in place, creating a simulated toe catch.
After the simulated perturbation (from $80\%$ of the nominal $\theta$-range until heel strike), the human subsystem attempted to complete the step.
For each step length, we created a range of $10$ final configurations, centered around the nominal final configuration, that widened or narrowed the base of support (BOS).

As a reminder, the general method presented in Sec. \ref{sec:theory} will eventually utilize able-bodied human stumble data to design the prosthetic knee trajectory library. 
However, in this implementation, we needed to simulate the stance knee behavior that would lead to a successful recovery.
For each combination of step length and final configuration, we designed stance knee behavior by specifying velocity values for the stance knee to track during and after the stumble.
These reference velocity values (scalars) were initially hand-picked for the smallest and largest step lengths, then interpolated across step lengths and final configurations to create $100$ successful recoveries.
We nondimensionalized time during and after the stumble (only until heel strike) by introducing a unitless time variable $\tau$.
Due to discontinuities in joint velocity after the sudden toe catch, the normalized time range was divided into two subsets: $\tau \in [0, 0.5]$ during the stumble, and $\tau \in (0.5, 1]$ after the stumble.
The joint velocities during and after the stumble were then normalized according to these new time ranges and denoted as $\dot{\tilde{q}}^N(\tau)$ for $\tau \in [0,1]$.
We then defined an additional variable corresponding to the system's recovery strategy, $k_j$, for $j = \{h,p\}$.
The parameter $k_{j}$ ranged from $[0, 1]$, with increasing value corresponding to a larger base of support at the final configuration.
We assume that for a given step length, each joint in the human subsystem recovers with the same desired final configuration, corresponding to $k_{h}$, but the prosthetic joint's trajectory parameter $k_p$ can be different from $\kh$.
Lastly, for each step length, we created parameterized velocity trajectories for joint $i$ in the $j^\text{th}$ subsystem by fitting polynomial functions of $\tau$ and $k_j$.
We used every $3^{\text{rd}}$ trial to fit the polynomial coefficients.
The polynomials, degree $4$ in $\tau$ and degree $1$ in $k_{j}$, are given by
\begin{equation}
\label{eq:param_velocities}
    \dot{\tilde{q}}^N_{j,i}(\tau ; k_{j}) = \begin{cases} a_{j,i}^1 + a_{j,i}^2 k_{j} + a_{j,i}^3\tau + ... + a_{j,i}^6\tau^4, & \tau \in [0, 0.5] \\
    b_{j,i}^1 + b_{j,i}^2 k_{j} + b_{j,i}^3\tau + ... + b_{j,i}^6\tau^4, & \tau \in (0.5, 1].
\end{cases}
\end{equation}

\subsection{Target Set}
\label{subsec:implementation_target}
As detailed in Sec. \ref{subsec:theory_target}, we defined the target set $\XT$ as the set of states at the end of the stumble step that would allow the system to converge to a nominal gait.
In this implementation, we formulate $\XT$ in terms of the center of mass (COM) positions, COM velocities, and swing foot positions at heel strike.
Although we define $\XT$ in terms of the model states themselves in Def. \ref{def:target_set}, here we choose a simpler target set which we find sufficient to describe the boundary between successful and unsuccessful trajectories.

\begin{figure}
    \centering
    \includegraphics[width=\columnwidth]{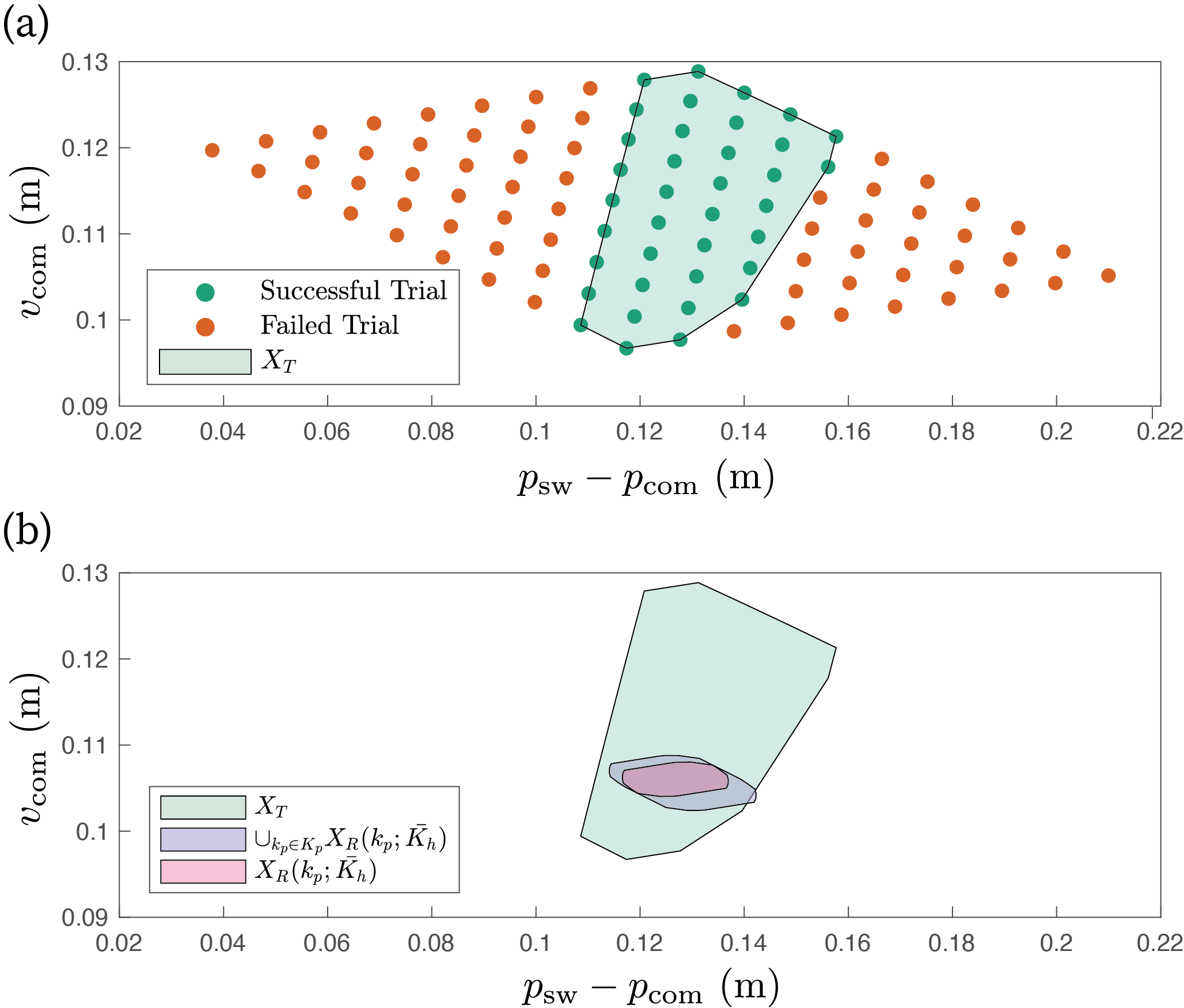}
    \caption{We used sampled polynomial-fit trials to form a target set for each step length. 
    (a) shows the normalized horizontal COM velocity and relative COM position for successful (green) and failed (orange) trials for a specific step length at $\tau = 1$.
    The polytope target set (shaded green) is constructed by the convex hull of the successful trials.
    (b) shows the same polytope target set along with the system's reachable sets described in Section \ref{sec:implementation:rotatotopes}.
    The blue shaded region represents the union of reachable sets for $\kp \in \Kp$, given some $\Khbar$.
    Because it is not completely contained within the target set $\XT$, there are unsafe $\kp \in \Kunsafe(\Khbar)$.
    A particular choice of $\kp$ yields the reachable set shown as the pink shaded region, which is considered safe because it is completely contained within $\XT$.
    Note, velocities are given in meters because time has been nondimensionalized.}
    \label{fig:target_set}
\end{figure}

The target sets are step-length specific, corresponding to the $10$ nominal gaits.
For each step length, we sampled combinations of $\kh$ and $\kp$, evaluated their associated polynomial-fit velocity and position trajectories, and simulated nominal walking from the system's state at $\tau = 1$.
These trials were marked as successes or failures based on whether they converged to the nominal gait.
Plots of the system's horizontal COM velocity at $\tau = 1$ versus the relative COM position (the horizontal distance from the COM position to the swing foot position at heel strike) revealed a sampled boundary between successful and failed trials. 
We formed $\XT$ by constructing polytopes defined by the convex hull of each step length's successful sampled trials, shown in Fig. \ref{fig:target_set}.

\subsection{Reachable Sets of the Human-Prosthesis System} 
\label{sec:implementation:rotatotopes}

Although in \eqref{eqn:FRS_theory} we write the reachable sets in terms of the full state space, here we use relative COM positions and COM velocites to match the target set.
TRIP-RTD utilizes parameterized reachable sets of these variables in an online optimization.
Computing these reachable sets directly is challenging; the reachable sets depend on the initial condition at the time the stumble occurs, and the trajectory parameterization is high-dimensional and nonlinear in time.
A recently developed method, ARMTD \cite{holmes20}, addresses these challenges by composing the reachable set of a multi-link robot's volume in workspace from reachable sets of each joint, allowing each joint's initial condition to be quickly incorporated.

Within ARMTD, reachable sets for each joint representing sets of rotations are computed offline using the trajectory parameterization \eqref{eq:param_velocities} and the open-source MATLAB toolbox CORA \cite{althoff2015}.
Online, ARMTD modifies these sets by the initial configuration at the start of the stumble $q_0$.
It uses them to compute parameterized reachable sets for each joint position in workspace relative to a predecessor joint.
Then, ARMTD ``stacks'' these sets of positions (via a Minkowski sum) according to the kinematic tree structure of the model.

Given $\Khbar$, we utilize ARMTD to compute the reachable sets of horizontal swing foot positions, COM positions, and COM velocities at heel strike, which we denote:
\begin{equation}
    X_i( \cdot \, ; \Khbar): \Kp \to \P(\R), \quad i = \{ p_\regtext{sw}, p_\regtext{com}, v_\regtext{com} \}.
\end{equation}
While $\Rsw$ is computed directly by ARMTD (\cite[Alg. 2]{holmes20}), we obtain $\Rcomp$ and $\Rcomv$ by modifying the stacking process according to the mass of each link.
Then, we obtain $\FRS(\cdot \, ; \Khbar): \Kp \to \P(\R^2)$ as
\begin{equation}
\scalebox{0.85}{
    $\FRS(\kp; \Khbar) = \left( \Rsw(\kp; \Khbar) \oplus -\Rcomp(\kp; \Khbar) \right) \times \left( \Rcomv(\kp ; \Khbar) \right)$
    }
\end{equation}
where $\oplus$ represents a Minkowski sum.
Figure \ref{fig:target_set} shows $\FRS(\kp ; \Khbar)$ relative to $\XT$ for a particular choice of $\kp$.
Importantly, $\FRS$ produces an overapproximation of the true reachable set \cite[Lem. 7, 9]{holmes20}, which ensures safety within an optimization problem described in the next section.

\subsection{Online Trajectory Optimization}
\label{sec:implementation:online}
The last step of TRIP-RTD solves a constrained optimization problem for an optimal trajectory parameter $\kpopt$ which guarantees the system ends in $\XT$ for a range of human trajectory parameters $\Khbar \subseteq \Kh$. 
We use the target set $\XT$ and reachable set $\FRS$ to generate the constraint for this problem, which we show conservatively overapproximates the set of unsafe prosthetic trajectory parameters $\Kunsafe(\Khbar)$.

\begin{lem}
\label{lem:constraint}
Given a subset of plausible human trajectory parameters $\Khbar \subseteq \Kh$, there exists a function $c(\cdot \, ; \Khbar): \Kp \to \R^{n_c}$  whose zero superlevel set overapproximates the set of unsafe trajectory parameters:
\begin{equation}
    \Kunsafe(\Khbar) \subseteq \{ \kp \  | \ c(\kp; \Khbar) \geq 0 \}
\end{equation}
where the inequality is understood element-wise.
In particular, $c$ can be written as a linear function composed with a polynomial function of $\kp$, given by
\begin{equation}
    \label{eq:constraint_definition}
    c(\kp; \Khbar) = A\XRpoly(\kp; \Khbar) - b
\end{equation}
where $A \in \R^{n_c} \times \R^2$, $\XRpoly(\cdot \, ; \Khbar): \Kp \to \R^2$, and $b \in \R^{n_c}$ all depend on $\Khbar$.
\end{lem}
\begin{proof}
Due to space considerations, this proof refers extensively to \cite{holmes20}.
As in \cite[Equation 19]{holmes20}, the output of $\FRS$ can be divided into two sets:
\begin{equation}
    \FRS(\kp; \Khbar) = \XRbuf(\Khbar) \oplus \{ \XRpoly(\kp; \Khbar) \}
\end{equation}
The first set $\XRbuf(\Khbar) \subset \R^2$ is a polytope that is independent of $\kp$ (see \cite[Lemma 9]{holmes20}) and will ``buffer'' the target set $\XT$.
The point $\XRpoly(\kp ; \Khbar)$ has the property that it can be written as the evaluation of a polynomial $\XRpoly(\cdot \, ; \Khbar): \Kp \to R^2$, which is generated as in \cite[Equation 20]{holmes20}.

Next, because $\XRbuf(\Khbar)$ is a subset of $\FRS(\kp ; \Khbar)$ that is independent of $\kp$, we use it to conservatively buffer (i.e., reduce the size of) $\XT$.
To do so, we take the Minkowski (or Pontryagin) difference \cite{herceg2013} of the polytopes $\XT$ and $\XRbuf(\Khbar)$ to form a \defemph{constraint polytope} $X_C$:
\begin{equation}
    X_C = \{ \tilde{x} \in \R^2 \ | \ \tilde{x} \oplus \XRbuf(\Khbar) \subseteq \XT \}
\end{equation}
We note that $X_C$ may not always be non-empty, for example, if $\XRbuf(\Khbar)$ is very large.
In this case, we cannot verify the safety of any $\kp$, and let $c$ be $+\infty$.
If $X_C$ is non-empty, we proceed and let $(A,b)$ denote the half-space representation of the constraint polytope, so $X_C = \{ \tilde{x} \in \R^2 \ | \ A\tilde{x} - b < 0 \}$.
The Minkowski difference and half-space representation for $X_C$ are computed using the MPT3 MATLAB toolbox \cite{herceg2013}. 

For a given $\kp$, $\XRpoly(\kp ; \Khbar) \in X_C \implies \FRS(\kp ; \Khbar) \subseteq \XT$, which follows from the definition of $X_C$.
Therefore, we have
\begin{equation}
    A\XRpoly(\kp; \Khbar) - b < 0 \implies \FRS(\kp; \Khbar) \subseteq \XT.
\end{equation}

Choosing $c$ as in \eqref{eq:constraint_definition}, $c(\kp; \Khbar) < 0$ ensures that the reachable set $\FRS(\kp; \Khbar)$ is contained in the target set $\XT$.
The superlevel set $\{ \kp \  | \ c(\kp; \Khbar) \geq 0 \}$ is not verified, and therefore overapproximates $\Kunsafe(\Khbar)$.
\end{proof}

Finally, we solve a constrained optimization problem to find the best \emph{safe} trajectory parameter $\kp$ given $\Khbar$:
\begin{equation}
\label{eq:online_planning_opt}
    \kpopt = \regtext{argmin}_{\kp \in \Kp}\big\{\costfunc(\kp)\ |\ c(\kp; \Khbar) < 0 \big\}.
\end{equation}

\begin{thm}
\label{thm:feasible}
Any feasible solution to \eqref{eq:online_planning_opt} parameterizes a stumble-recovery trajectory of the powered prosthesis that will converge to the nominal walking trajectory $\xnom$ for all $\kh \in \Khbar$.
\end{thm}
\begin{proof}
Lemma \ref{lem:constraint} ensures $\FRS(\kp ; \Khbar) \subset \XT$ for any feasible $\kp \in \Kp$.
By Def. \ref{def:target_set}, $\XT$ is a subset of the basin of attraction of the nominal walking trajectory $\xnom$ under the nominal prosthesis controller $\unom$.
\end{proof}

Though \eqref{eq:online_planning_opt} is a nonlinear program, the structure of $c$ makes it easy to solve in practice.
Since $c$ is a linear function of a polynomial of $\kp$, we derive analytic gradients of $c$ with respect to $\kp$.
We solve \eqref{eq:online_planning_opt} using MATLAB's \texttt{fmincon}, with $c$ composed of the polytope half-space representation, $(A, b)$ computed using MPT3 \cite{herceg2013}, and the polynomial $\XRpoly(\cdot \, ; \Kh)$ found using ARMTD \cite{holmes20}.
When the optimization is infeasible, for example if $\XT$ is too small or $\Khbar$ is too large, we can make no claims about any $\kp$.

We note that Thm. \ref{thm:feasible} ensures any $\kp$ feasible to \eqref{eq:online_planning_opt} produces a \defemph{desired trajectory} which results in stable walking.
However, we make no guarantee that the prosthetic subsystem achieves this desired trajectory.
To experimentally show the efficacy of TRIP-RTD, in the next section we design a feedback controller to track the trajectory parameterized by $\kpopt$, and test stumble recovery success across varying step lengths and subsets $\Khbar$.
Note that this validation is done using the \defemph{full dynamic model} of the human-prosthesis system.

\section{Simulated Experiment}
\label{sec:experiment}
\begin{figure*}
\centering
\includegraphics[width=\textwidth]{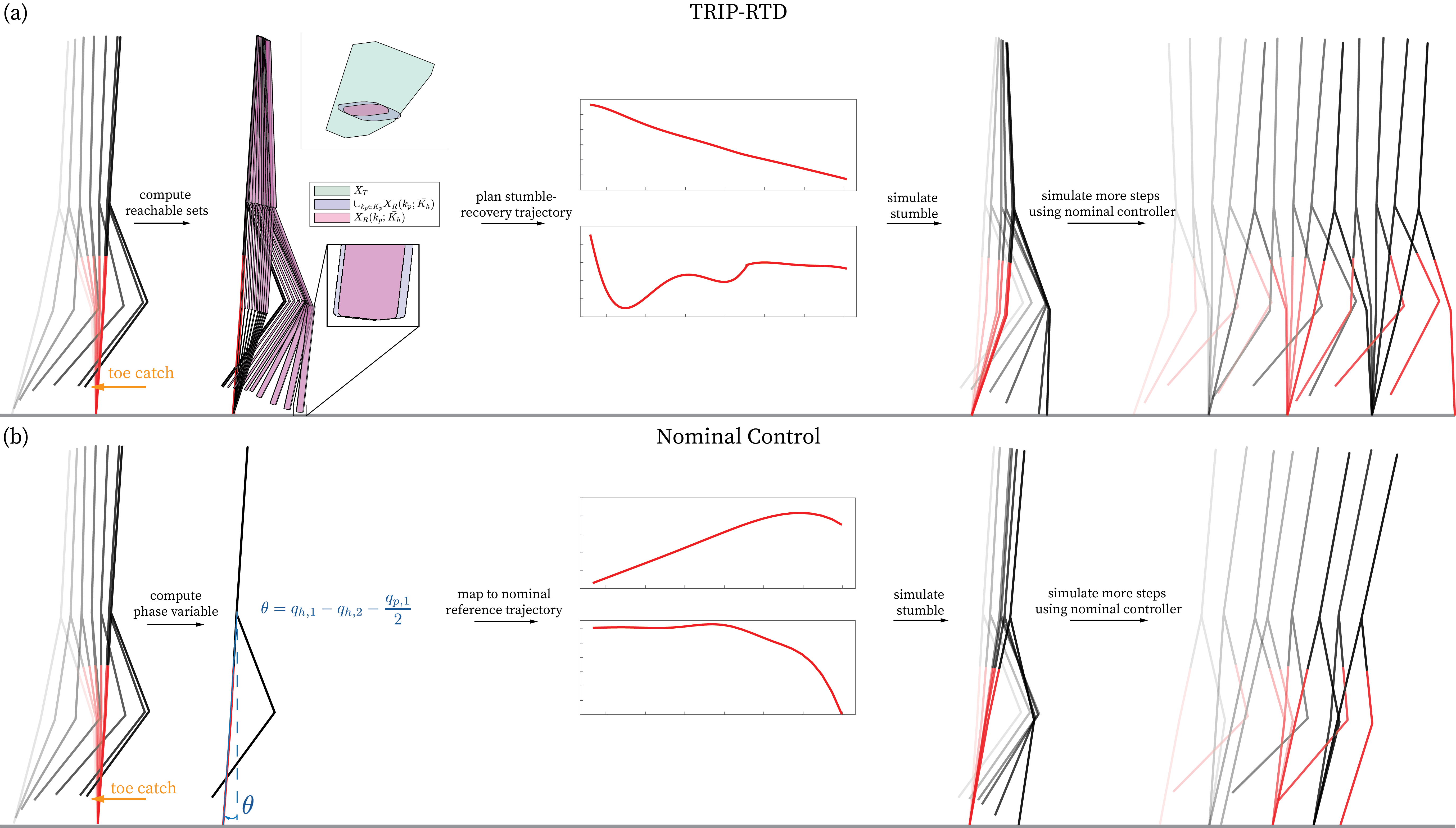}
\caption{We simulated the same toe-catch stumble using two different reference trajectories for the prosthetic knee. 
(a) shows the reference trajectory generated by TRIP-RTD, which selected a $\kp$-value that ensured a successful recovery for all $\kh \in \Khbar$. 
When tracking this trajectory, the system was able to return to a nominal walking gait after the stumble. 
(b) shows the same stumble while the prosthetic knee tracked a nominal trajectory using phase-based control, with the stance leg angle as the phase variable. 
The system was unable to converge to a nominal gait after the stumble, and subsequently fell over.}
\label{fig:experiment_results}
\end{figure*}
This section describes our assessment of the proposed method using simulated human stumble-recovery data.
Recall that for each nominal step length, the system's stumbles/recoveries can vary according to the $\kh$ and $\kp$ parameters, which range from $0$ to $1$.
Offline, we divided $\Kh = [0, 1]$ and $\Kp = [0, 1]$ into $10$ uniform subintervals of $\Khbar \subset \Kh$ and $\Kpbar \subset \Kp$, and computed step-length-specific reachable sets for each joint within these ranges.
Because the conservatism of the reachable sets is proportional to the size of $\Kpbar$ \cite{holmes20}, for each step length and $\Khbar$ we solve $10$ optimization problems \eqref{eq:online_planning_opt}, one for each $\Kpbar$.
We use a cost function of $\costfunc(\kp) = 0$ to test that any $\kp$ \defemph{feasible} to the constraints will be successful.
Online, we create the associated reachable sets for a particular step length and subintervals $\Khbar$ and $\Kpbar$, and solve (\ref{eq:online_planning_opt}) for $\kpopt \in \Kpbar$ that ensures $\FRS(\kpopt; \Khbar)$ is contained within $\XT$.
Then, out of these $10$ possible $\kpopt$, we choose any feasible $\kpopt$ for testing in simulation.
In practice, these optimizations can be performed in parallel, though we performed them in sequence here.
To demonstrate the capability of TRIP-RTD for real-time optimization, we enforced a timeout of $75$ms for each optimization within which a feasible $\kpopt$ must be found.
We repeated this process for each step length and $\Khbar$.

We next evaluated (\ref{eq:param_velocities}) for each $\kpopt$ solution to create a prosthetic knee trajectory, $\tilde{q}^N_{p,1}(\tau; \kpopt)$ and $\dot{\tilde{q}}^N_{p,1}(\tau; \kpopt)$, for $\tau \in [0,1]$, to use for a specific step length and $\Khbar$.
We converted time back from dimensionless $\tau$ to $t$, scaled the velocity trajectory accordingly, and parameterized the trajectories by $\theta$ for use in simulation.
Next, we randomly sampled three values of $\kh$ from $\Khbar$, corresponding to different human stumble recoveries.
We simulated these three stumbles two separate times to compare the TRIP-RTD knee trajectory to the system's nominal controller detailed in Fig. \ref{fig:experiment_results}.
First, we simulated the stumbles while the prosthesis used PD control to track the TRIP-RTD trajectory as described in \eqref{eq:feedback}.
Next, we simulated the same stumbles while the prosthesis used PD control to track the nominal stance-knee trajectory associated with the step length.
In both tests, the entire system switched back to nominal phase-based control after the stumble step heel strike, with each joint using PD control to track the nominal trajectories.
The trials were labeled as successes or failures based on whether they converged to the nominal gait after the stumble step or fell over.
\section{Results}
\label{sec:results}
We tested TRIP-RTD over $10$ different step lengths, each with $10$ subintervals of $\Khbar$.
Of the $100$ $\Khbar$-subintervals we optimized over, $71$ returned at least one feasible $\kpopt$-value in the allotted time.
For each $\Khbar$ with a feasible $\kpopt$, we ran the simulated experiment described in Section \ref{sec:experiment}.
We tested three sampled stumbles per $\Khbar$ with feasible $\kpopt$, totaling $213$ tests.
The nominal knee controller resulted in failure for every trial, while the stumbles simulated with TRIP-RTD reference trajectories were $100\%$ successful.
Examples of successful (TRIP-RTD) and failed (nominal phase-based) stumble recoveries are shown in Fig. \ref{fig:experiment_results}.

The mean and max online planning time for each trial that returned a feasible $\kpopt$ was $69.7$ and $101$ ms, respectively. 
These times include computing the reachable sets associated with $\Khbar$ and $\Kpbar$, generating the target set constraint, and solving (\ref{eq:online_planning_opt}).
\section{Discussion}
\label{sec:discussion}
Prosthetic stumble-recovery controllers could help people with lower-limb amputations perform everyday tasks with more confidence.
However, the uncertainty in measuring and estimating human behavior have made designing these controllers difficult.
In this paper, we presented a method for planning stumble-recovery trajectories online while explicitly accounting for a range of predicted human responses.
Our method constructs reachable sets that depend on both the human and prosthesis subsystems, then selects a trajectory parameter for the prosthetic knee online to ensure that the entire system recovers successfully.
To our knowledge, this work represents the first prosthetic controller framework that can plan trajectories online while accounting for a range of possible behaviors.

The majority of simulated stumbles returned a feasible $\kpopt$ and associated successful recovery trajectory, but infeasible stumbles occurred at the lowest and highest $\Khbar$ subsets (corresponding to the narrowest and widest base of support) for each step length.
These cases lived on the edge of the target set, and thus were more susceptible to the system's overapproximative reachable sets exiting the target set, even for a $k_p$-value that could have led to a successful recovery.
The current framework also assumes that arriving in a target set defined by the system's COM relative position and velocity is sufficient for the system's return to nominal behavior.
However, the TRIP-RTD framework can include other notions of target sets, informed by human stumble experiments.

Although this paper describes an implementation using simulated human stumbles, the method is generalizable.
The theoretical approach outlined in Sec. \ref{sec:theory} can be applied to walking models with different dimensions, parameters and dynamics.
The method could be applied to other types of trips or slips, regardless of whether the prosthesis is in stance or swing, and the cost can include goals such as minimizing the work done by the prosthesis.
Lastly, the stumble-recovery behavior could be parameterized by different physical quantities, such as the common stumble-recovery swing-leg behaviors detailed in \cite{shirota2014}.
Data from human stumble-recovery experiments can be used to create parameterized trajectories and reachable sets.
Our future work will generalize TRIP-RTD in each of these areas with the goal of incorporating this framework into prosthetic hardware.
The simulation results represent a promising step in developing prosthesis controllers that can increase confidence and safety by working alongside a human to recover from stumbles.

\renewcommand{\bibfont}{\normalfont\footnotesize}
{\renewcommand{\markboth}[2]{}
\printbibliography}

\end{document}